\documentclass[letterpaper, 10 pt, conference]{ieeeconf}  

\IEEEoverridecommandlockouts                              

\usepackage{multicol}
\usepackage[breaklinks,colorlinks,linkcolor=black,bookmarks=true]{hyperref}
\usepackage{amsmath}
\usepackage{amssymb}
\usepackage{graphicx}
\usepackage{booktabs}
\usepackage{subcaption}
\usepackage{multirow}
\usepackage{color}
\usepackage[font=small]{caption} %
\usepackage[T1]{fontenc}
\usepackage{xspace}
\usepackage{dblfloatfix}
\usepackage{placeins} 
\usepackage{booktabs}
\usepackage{multirow}
\usepackage{pifont}
\usepackage{cite}

\newcommand{\method}{\mbox{{PRISM}}\xspace}

\newcommand{\yes}{\color{blue}{\ding{51}}}
\newcommand{\no}{\color{red}{\ding{55}}}

\title{\LARGE \bf
PRISM: Precision and contact-rich Real-world Industrial Skill dataset with Multimodal sensing 
}

\author{Tengbo Yu$^{1,2}$\quad{}Jiahao Wu$^{2}$\quad{}Hanning Wang$^{1}$\quad{}Rui Chen$^{3}$\quad{}Chuanhou Liu$^{4}$\quad{}Chuang Sun$^{5}$\quad{}Hangxin Liu$^{1\dagger}$
\thanks{$^\dagger$ Corresponding author. Email: \tt{hx.liu@pku.edu.cn}.}
\thanks{$^{1}$ State Key Laboratory of General Artificial Intelligence, School of Intelligence Science and Technology, Peking University, Emails: {\tt\small yutengbo26@stu.pku.edu.cn}. $^{2}$ Delta Intelligence. $^{3}$ PKU-Wuhan Institute for Artificial Intelligence. $^{4}$ Hubei Humanoid Robot Innovation Center Co., Ltd. $^{5}$ China Academy of Information and Communications Technology.}
}

\begin{document}

\maketitle

\begin{abstract}
Recent progress in robotic learning has been fueled by large-scale datasets collected in everyday environments. However, most existing datasets emphasize short-horizon, low-contact tasks such as pick-and-place, and therefore do not capture the precision control, force/torque or tactile regulation, and multimodal feedback required for industrial assembly. To address this gap, we introduce \method, a large-scale multimodal dataset for contact-rich industrial operations. The dataset spans more than 25 manipulation tasks (e.g., \textit{electronic components plug/unplug, conveyor-based sorting}) and covers diverse mechanical constraints. \method includes more than 5,000 trajectories totaling 45 hours of teleoperated demonstrations, recorded using synchronized multi-view RGB-D, force/torque, tactile, and robot-state measurements. In contrast to datasets collected in household or laboratory settings, \method provides a realistic benchmark for multimodal perception and control under high-precision industrial constraints, and serves as a foundation for contact-rich, generalizable manipulation in real-world manufacturing environments. The dataset is open-sourced at: \url{https://tengbo-yu.github.io/PRISM/}
\end{abstract}

\section{Introduction}
\label{sec:intro}

\begin{figure*}[!t]
    \centering
    \includegraphics[width=1\textwidth]{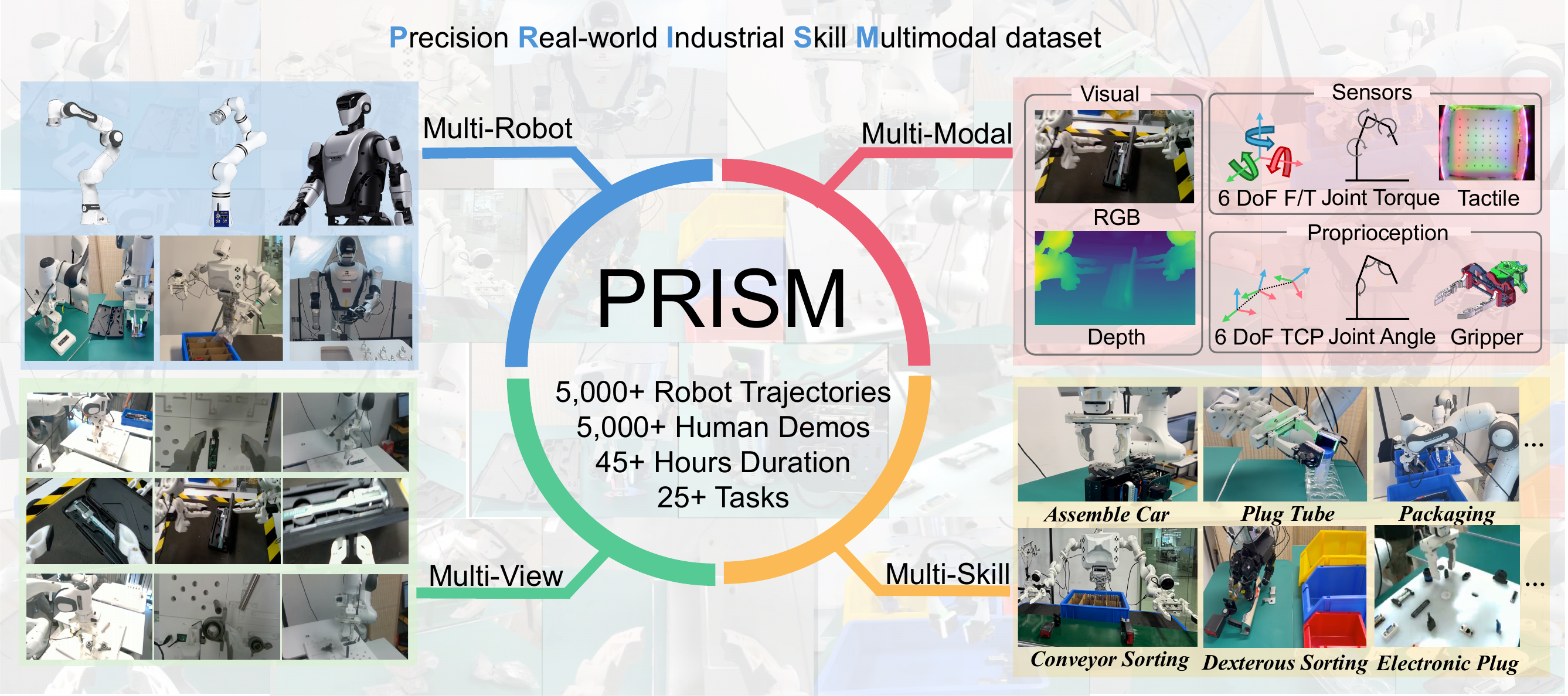}
    \caption{ \textbf{Overview of \method dataset.} We collect data using multiple robot platforms in diverse environments. Each manipulation episode includes synchronized multimodal observations, including visual, force, and proprioceptive signals, with tactile signals available for a subset of episodes.. For each episode, the full manipulation process is captured with well-calibrated multi-view cameras. The dataset spans a wide range of manipulation skills, and each episode is paired with a corresponding language description. In total, we provide more than 5,000 robot trajectories and 5,000 paired human demonstrations, covering over 45 hours of interaction across more than 25 tasks.
    }
    \label{fig:teaser}
\end{figure*}

Robotic learning has recently undergone a paradigm shift driven by large-scale datasets collected in diverse everyday environments \cite{Yaprcr2019deeplearning}. By leveraging broad visual coverage and imitation learning at scale, robots have demonstrated encouraging generalization for short-horizon, low-contact manipulation skills \cite{lu2025anybimanual,yu2025manigaussian++,black2026pi0visionlanguageactionflowmodel,kim2025fine}. Such datasets have powered substantial advances in modern robotics, allowing learning-based methods to perform strongly on a wide range of problems, from motion planning \cite{chi2024diffusionpolicy,kim2024openvla,brohan2022rt, zitkovich2023rt} and task planning \cite{ahn2022can,huang2022inner, rana2023sayplan, driess2023palm} to locomotion \cite{RoboImitationPeng20,radosavovic2024real} and anomaly detection \cite{ahn2022can, Sinha-RSS-24,liu2023reflect}. However, this progress has not yet translated to industrial assembly, where task success depends on tight tolerance, persistent contact, and precise force/torque or tactile regulation. In such settings, vision alone is often insufficient: minute misalignments, frictional sticking, compliance, and insertion jamming can be visually ambiguous but are immediately expressed through force/torque   or tactile signals and other contact cues \cite{morgan2021vision, zhao2024tac, garcia2009survey}. Consequently, industrial manipulation demands learning not only from perception, but also how to control interaction reliably throughout long-horizon procedures.


A key bottleneck is the mismatch between existing robot learning datasets and the requirements of industrial operations. Many widely used datasets emphasize primitive skills (e.g., pick-and-place, in-hand manipulation, lifting, and stacking) \cite{o2024open, shafiullah2023bringing, khazatsky2024droid} that are typically short-horizon and weakly contact-dependent \cite{fang2023rh20t, khazatsky2024droid, o2024open}, and therefore do not provide the high-resolution contact information needed for precision assembly. As highlighted in prior work, such primitive-focused datasets "failed to provide the critical data necessary for precise, contact-rich manipulation," while long-horizon assembly and disassembly require accurate contact measurements—especially force/torque and tactile—that visual sensing alone cannot reliably capture \cite{wu2024tacdiffusion, huang2025tactile, zhang2025vtla,zhang2026craft}. Recent efforts, such as REASSEMBLE, begin to address long-horizon and contact-rich scenarios by incorporating richer sensing streams such as multi-view cameras and six-axis force/torque measurements, demonstrating the importance of multimodal supervision for tight-tolerance manipulation \cite{sliwowski2025reassemble, yu2025forcevla}. However, the remaining gap is not only dataset scale. Existing resources rarely combine multi-embodiment robot platforms, synchronized multimodal sensing, industrial contact-rich procedures, and diverse teleoperation interfaces within one unified dataset. This combination is essential for learning policies that must transfer across robot bodies, sensing modalities, operators, and contact conditions in production-like settings.

To bridge this gap, we present \method, a large-scale multimodal dataset for contact-rich industrial operations. \method covers 25 manipulation tasks, including \texttt{installing bearings, automotive material sorting} and so on, spanning diverse mechanical constraints representative of real industrial processes. We collect over 5,000 teleoperated trajectories totaling more than 45 hours, recorded with synchronized multi-view RGB-D, six-axis force/torque, and robot state streams, with tactile sensing included for a subset of episodes. We provide a comparison of commonly used robot learning datasets and their properties in Table~\ref{tab:datasets}. Compared with existing datasets collected in household or laboratory environments, \method offers a realistic benchmark for learning multimodal policies under high-precision industrial constraints, emphasizing contact-rich interactions and stringent tolerances. Crucially, \method is designed around three dataset principles aligned with industrial requirements: higher capture standards, broader multimodal coverage, and large-scale diversity. First, we emphasize high-fidelity sensing, accurate calibration, and strict temporal synchronization to ensure that subtle contact events and force/torque transitions, as well as tactile transitions when available, are captured reliably. Second, we provide comprehensive multimodal observations that support learning contact-rich representations and controllers beyond purely visual policies. Third, we scale demonstrations across multiple tasks, robots, data collection platforms, and constraint structures to encourage generalization across industrial operation families. By offering a rich, multimodal dataset, \method fosters the development of adaptive and versatile robotic systems capable of tackling the challenges of long-horizon, contact-rich manipulation. To summarize, our contributions are as follows:
\begin{enumerate}
    \item \textbf{Industrial realism with contact-rich constraints.} We introduce \method, a multimodal dataset centered on high-precision industrial operations with diverse mechanical constraints.
    \item \textbf{High-fidelity synchronized multimodal data.} \method provides time-aligned multi-view RGB-D, six-axis force/torque, robot state streams, and tactile signals for a subset of episodes for contact-rich learning.
    \item \textbf{Scale for generalizable industrial manipulation.} \method contains more than 5,000 trajectories and more than 45 hours of demonstrations, enabling systematic training and evaluation of multimodal perception and control under industrial constraints.
\end{enumerate}


\section{Related Work}
\label{sec:related_work}


\begin{table*}[t]
    \centering
    \begin{tabular}{lcp{3cm}p{2.5cm}p{3cm}c}
    \toprule
    Dataset & \# Demos & Sensors & Robot & Collection Method & Contact Rich \\
    \midrule
    BC-Z~\cite{jang2022bc} & 263k & 1 RGB Camera, Robot Proprioception & Everyday Robots & VR teleoperation & \no \\
    RT-1~\cite{brohan2022rt} & 130k & 1 RGB Camera, Robot Proprioception & Everyday Robots & VR teleoperation & \no \\
    BridgeDatav2~\cite{walke2023bridgedata} & 60.1k & 4 RGB Cameras, 1 Depth Camera, Robot Proprioception & WidowX & VR teleoperation & \no \\
    FurnitureBench~\cite{heo2025furniturebench} & 5.1k & 2 RGB Cameras, Robot Proprioception & Franka & VR teleoperation & \yes \\
    DROID~\cite{khazatsky2024droid} & 76k  & 3 RGBD Cameras, Robot Proprioception & Franka Emika Panda Research 3 & VR teleoperation & \yes \\
    RH20T~\cite{fang2023rh20t} & 110k &  8-10 RGBD Cameras, 1 Microphone, F/T Sensor & Multiple & haptic teleoperation & \yes \\
    Assembly101~\cite{sener2022assembly101} & 4.3k & 8 RGB Cameras, 4 Mono Cameras & N/A & Human demonstration & \yes \\
    FAILURE~\cite{inceoglu2021fino} & 229 & 1 RGBD Camera, 1 Microphone & Baxter & Scripted & \no \\
    REASSEMBLE~\cite{sliwowski2025reassemble} & 4k & 3 RGB cameras, Robot Proprioception, 1 Event camera, 3 Microphones, F/T Sensor & Franka Emika Panda Research 3 & Haptic teleoperation & \yes \\
    \midrule
    \textbf{PRISM} & 5k & 2-4 RGBD Cameras, Robot Proprioception, F/T Sensors, \textbf{Tactile Sensors} & Multiple robots with \textbf{3 different types end-effector} & \textbf{Exoskeleton, Tracker, VR teleoperation} & \yes \\
    \bottomrule
    \end{tabular}
    \caption{\textbf{Datasets comparison.} We compare several commonly used datasets based on the number of demonstrations, the sensors used during data collection, the robotic platform, the data collection method, and whether the dataset contains contact-rich tasks.}
    \label{tab:datasets}
\end{table*}

\subsection{Large-Scale Datasets in Robotic Manipulation}
In recent years, the field of robot learning has undergone a significant paradigm shift, moving from small-scale, task-specific datasets \cite{lenz2015deep,mahler2017dex} toward large-scale, cross-embodiment data \cite{dasari2019robonet,brohan2022rt,walke2023bridgedata}. To achieve stronger generalization and train universal robot foundation models, the community has introduced several milestone datasets by integrating data from various robotic platforms.

For instance, Open X-Embodiment~\cite{o2024open} aggregates hundreds of thousands of demonstration trajectories, aiming to cover diverse environments and tasks through sheer data volume. Research such as RT-1~\cite{brohan2022rt} and BridgeData V2~\cite{walke2023bridgedata} has further demonstrated that models trained on massive datasets can exhibit impressive zero-shot transfer capabilities, adapting to unseen instructions and scenarios. The success of these datasets is primarily attributed to their scale, which enables models to learn general visuomotor priors. This allows them to perform object manipulation and relocation tasks within unstructured, real-world settings.

\subsection{High-Precision Datasets for Industrial Manipulation}
Recent advancements in robot learning have been driven by the emergence of large-scale datasets collected across diverse environments. Foundation initiatives such as Open X-Embodiment~\cite{o2024open}, RT-1~\cite{brohan2022rt}, and BridgeData V2~\cite{walke2023bridgedata} have aggregated extensive demonstrations to enable general-purpose robotic control. These datasets typically focus on unstructured settings, such as household or kitchen environments, where robots perform fundamental, short-horizon tasks like picking, placing, and object rearrangement.

To address the limitations of primitive skills, research has shifted towards long-horizon and contact-rich manipulation tasks. FurnitureBench~\cite{heo2025furniturebench} has made progress in addressing long-horizon tasks. However, despite the extended temporal horizon, furniture assembly generally involves loose mating parts, lacking the tight tolerances of precision manufacturing.

To tackle high-precision challenges, recent benchmarks like REASSEMBLE~\cite{sliwowski2025reassemble} and RH20T~\cite{fang2023rh20t} have adopted standardized industrial protocols, such as the NIST Task Board. These works emphasize that visual sensing alone is often insufficient for tasks like gear meshing or connector insertion, necessitating the use of high-frequency force-torque sensors and proprioception. While incorporating such modalities improves physical feedback, it introduces hardware dependencies and complicates data collection, often limiting dataset scale (e.g., REASSEMBLE contains approximately 4,000 demonstrations).

In contrast, our work bridges the gap between large-scale generalist datasets and high-precision industrial needs. We posit that vision-based policies can implicitly learn the contact dynamics required for tight-tolerance assembly when supported by substantial data scale and hardware diversity. To this end, we introduce a dataset comprising 5,000 trajectories covering a wide range of industrial task scenarios, collected across multiple robotic platforms including Franka, Realman, and LEJU. Furthermore, to enable rigorous evaluation of interaction quality, we incorporate a subset of data equipped with tactile sensors, serving as a high-fidelity benchmark for fine-grained testing and validation.

\begin{table}[t]
\centering
\begin{tabular}{llll}
\hline
Conf. & Robot & Gripper & Teleoperation \\
\hline
Cfg 1 & Franka & 3D Printed & Tracker\\
Cfg 2 & Franka & Visuotactile Gripper & Tracker \\
Cfg 3 & Franka & Visuotactile Dexterous Hand & Tracker \\
Cfg 4 & Franka & 3D Printed &  Exoskeleton\\
Cfg 5 & Realman & 3D Printed & Exoskeleton\\
Cfg 6 & Realman & 3D Printed &  VR\\
Cfg 7 & LEJU & Robotiq-85 & VR \\
\hline
\end{tabular}
\caption{Hardware specification of different configurations.}
\label{tab:hardware_spec}
\end{table}


\section{DATASETS}
\label{sec:datasets}

We introduce Precision and contact-rich Real-world Industrial Skill dataset with Multimodal sensing (PRISM), a dataset for contact-rich industrial manipulation. Fig.~\ref{fig:teaser} shows an overview, and Table~\ref{tab:datasets} compares \method with representative public datasets.

\subsection{Properties of \method}
\method is designed for general contact-rich industrial manipulation under high-precision constraints. We emphasize diversity, multimodal sensing, scale, and compositional task structure.

\begin{table}[h]
\centering
\begin{tabular}{lcc}
\toprule
Modal & Size & Frequency \\
\midrule
RGB image & $540 \times 960 \times 3$ & 15 Hz \\
Depth image & $540 \times 960$ & 15 Hz \\
Visuotactile image & $256 \times 256$ & 30 Hz \\
Robot joint angle & 6 / 7 & 15 Hz \\
Robot joint torque & 6 / 7 & 15 Hz \\
Gripper Cartesian pose & 6 / 7 & 15 Hz \\
Gripper width & 1 & 15 Hz \\
6DoF F/T & 6 & 100 Hz \\

\bottomrule
\end{tabular}
\caption{Data information of different modals.}
\label{tab:data_info}
\end{table}

\begin{figure*}[t]
	\centering
	\includegraphics[width=1\textwidth]{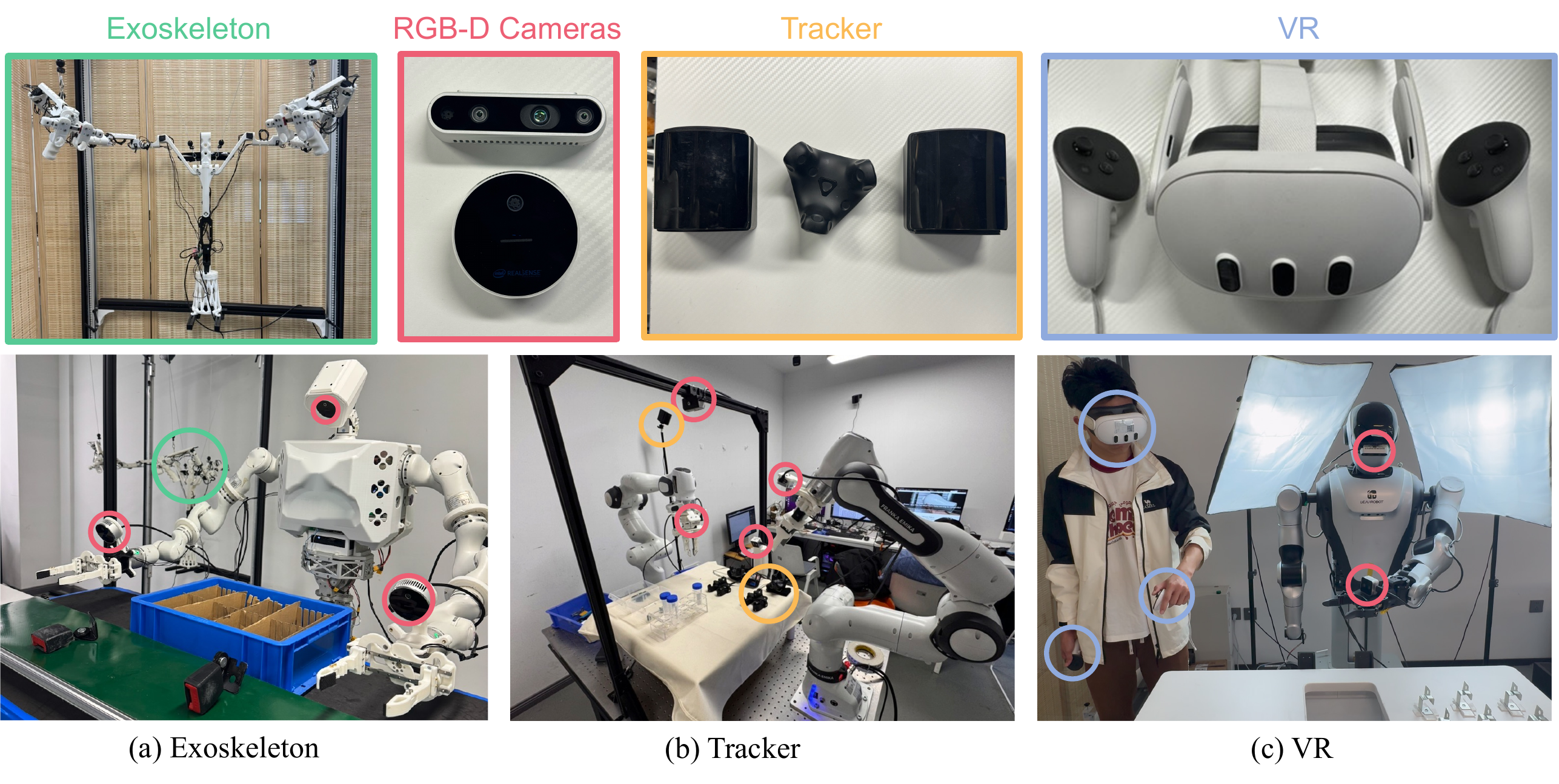}
	\caption{\textbf{Illustration of our data collection platform.} Including three different teleoperation platforms with multi-view RGB-D cameras.}
	\label{fig:teleoperation}
\end{figure*}

\textit{a)\ Diversity:} \
\method spans contact-rich tasks on the NIST Assembly Task Board \cite{kimble2022performance} and production-oriented tasks such as packaging, installation, and conveyor-based sorting. These tasks cover different mechanical constraints in industrial manipulation. We collect data with 3 robot platforms and 3 gripper types, allowing similar task families to appear under different kinematics, grasping affordances, and contact patterns. Hardware details are shown in Table~\ref{tab:hardware_spec}.

We further vary teleoperation interfaces, environments, and trajectories. Demonstrations are collected with 3 teleoperation modalities---exoskeleton, tracker, and VR-based control---as shown in Fig.~\ref{fig:teleoperation}. We change tabletop surfaces, backgrounds, lighting, object placements, and object poses across sessions. Data are collected by 8 volunteers, adding natural variation in motion style, contact timing, and corrective behaviors. We also include episodes with intentional human perturbations during manipulation to improve robustness to non-ideal contact events.

\textit{b)\ Multi-Modal:} \  
Each trajectory in \method is recorded with synchronized multimodal observations to capture both geometric perception and contact interaction signals that are critical for contact-rich industrial manipulation. Specifically, we log RGB-D images from the vision system to provide scene appearance and depth geometry, and visuotactile images from tactile sensors/end-effectors to directly observe local contact patterns, deformation, and slip cues. To characterize interaction dynamics, we record the 6-axis end-effector force/torque wrench as well as joint torques, enabling learning of force- and contact-sensitive control policies. In addition, we provide proprioception, which encompasses joint angles/torques, end-effector cartesian pose, and gripper states, offering a complete description of the robot's internal state for state-feedback control and multimodal sensor fusion. All information is collected at the highest frequency supported by our data collection platform and saved with corresponding timestamps, and the details are given in Table \ref{tab:data_info}

\begin{figure}[t]
	\centering
	\includegraphics[width=0.48\textwidth]{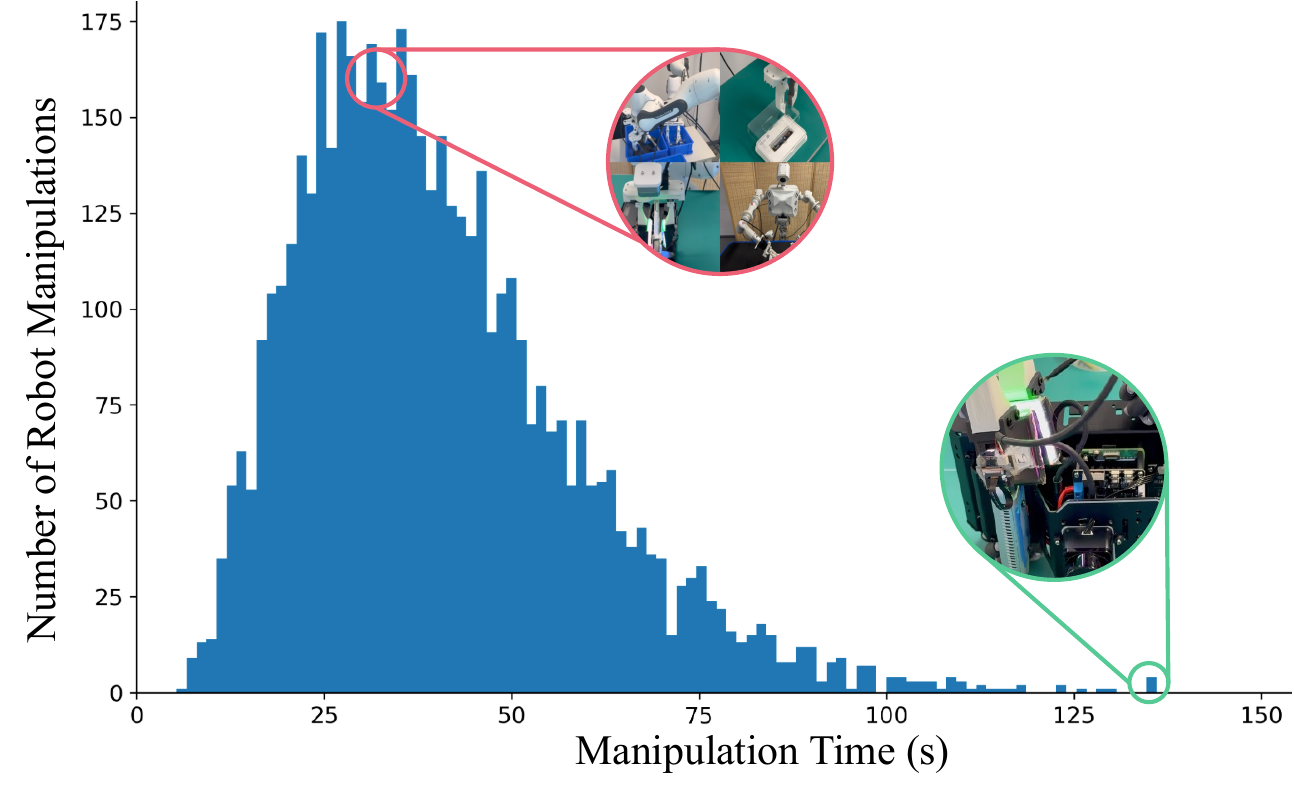}
	\caption{Statistics on the execution time of different robotic manipulations in our dataset.}
	\label{fig:statistics}
\end{figure}

\textit{c)\ Scale:} \ 
\method is collected at a substantial scale to support data-hungry multimodal learning. The dataset contains over 5,000 robot trajectories, together with an equal number of paired human demonstrations, totaling approximately 27 million images across vision and visuotactile streams. Fig. \ref{fig:statistics} summarizes the dataset statistics on the manipulation time for each episode in our dataset. Most episodes have durations ranging from 25 to 35 seconds, reflecting the sustained contact and multi-step nature of industrial operations. With its combination of large trajectory count, high-volume multimodal visual data and coverage across multiple industrial settings and robot embodiments, \method constitutes one of the most comprehensive multimodal datasets available for contact-rich manipulation in real-world industrial scenarios. 

\textit{d)\ Compositionality:} \ 
\method is organized to support learning at multiple stages by treating complex industrial procedures as compositions of reusable sub-tasks. Many real-world operations are naturally multi-step, where each step is meaningful on its own yet also contributes to a larger goal. For example, during the assembly of an autonomous patrol vehicle, installing the wheels, mounting the hubs, plugging the camera and installing the radar can be considered a single end-to-end task when viewed as the complete assembly process; at the same time, each step can be treated as an individual task with its own objectives, constraints and success criteria. This compositional structure allows users to (i) train models on fine-grained skills and evaluate them in isolation, (ii) study how performance scales with increasing procedural length, and (iii) combine learned components to solve longer-horizon industrial workflows under consistent sensing and annotation.

\subsection{Data Collection and Processing}

Unlike most existing datasets that rely on a single teleoperation interface, we collect demonstrations using three complementary teleoperation methods (Exoskeleton-, Tracker-, and VR-based). This design not only broadens the coverage of human control styles and interaction behaviors, but also enables a controlled comparison of how the same task, collected under different teleoperation modalities, differ in data quality (e.g., smoothness, precision, contact stability and failure rates) and how these differences translate into downstream learning performance. We further detail the hardware configurations, multimodal synchronization and calibration, and post-processing steps that ensure consistent, high-fidelity multimodal records across robots, end-effectors and collection sessions.

\textit{a)\ Collection:} \ 
Fig. \ref{fig:teleoperation} illustrates the three teleoperation platforms used to collect \method. (a) Exoskeleton teleoperation platform. This platform is a bimanual robot built with two Realman RM75-6F arms mounted on a torso with 3 waist DoFs, resulting in 17 DoFs for the full body, leading to a wider operation space. The system is equipped with 6-axis and end-effector force/torque sensors, two parallel-jaw grippers, a first-person head-mounted camera, and two wrist cameras. During data collection, we simultaneously record both the robot states and the exoskeleton joint angles, enabling paired human-robot motion traces for the same execution. (b) Tracker-based teleoperation platform. This platform consists of two Franka Emika Panda arms (14 DoFs total) with access to joint torque signals. Visual sensing includes two wrist cameras, a third-person camera, and an overhead camera. The end-effector is modular: the standard parallel jaw grippers can be replaced with either a visuotactile gripper or a visuotactile dexterous hand to capture contact-rich interactions with tactile images. (c) VR teleoperation platform. This platform uses a LEJU upper-body humanoid robot, instrumented with a wrist camera and a first-person head camera, and is controlled through a VR interface to collect demonstrations in scenarios that benefit from immersive first-person manipulation. We recruited 8 data-collection volunteers. Each volunteer underwent standardized training before collecting demonstrations. After collection, the same volunteers perform episode filtering, annotation, and scoring, providing structured quality control signals for downstream use.

\textit{b)\ Processing:} \ 
After data collection, we process raw logs into episode files with clean timestamps, calibrated sensor parameters, and metadata suitable for multimodal learning. Since different sensors run at different native rates, we keep original timestamps for all streams and provide per-episode indexing so that modalities can be aligned by timestamp during training (e.g., nearest-neighbor matching or interpolation). We remove incomplete episodes and trim leading/trailing idle segments based on task start/end markers and operator annotations.

\textbf{Calibration and frame unification.} For each platform, we compute and store the full transformation graph needed to express all observations in consistent coordinate frames. For the Realman exoskeleton-controlled bimanual platform (Fig. \ref{fig:teleoperation}-a), we unify the two arms, waist, head camera, and wrist cameras under a shared world frame; end-effector wrench measurements are expressed in a clearly defined sensor frame and optionally transformed to the end-effector or base frame. We also synchronize and align exoskeleton joint trajectories with the corresponding robot joint trajectories, storing both streams with consistent joint ordering and metadata so that paired human–robot kinematics can be directly compared. For the Panda tracker-based platform (Fig. \ref{fig:teleoperation}-b), we calibrate the wrist cameras, third-person camera, and top-down camera w.r.t. the robot base/world frame, and standardize joint torque conventions. When the end-effector is replaced by a visuotactile gripper or a visuotactile dexterous hand, we additionally calibrate the tactile camera(s) and record their intrinsics/extrinsics relative to the end-effector frame. For the LEJU VR platform (Fig. \ref{fig:teleoperation}-c), we calibrate the head and wrist cameras and align VR control signals with robot kinematics, yielding a consistent representation of end-effector motion and camera observations.

\textbf{Unified episode packaging.} Finally, we serialize all processed data into a common schema shared across the three platforms: robot states/actions (including joint angles/torques, end-effector pose, gripper states), end-effector wrench (when available), multi-view RGB-D, visuotactile imagery (when available), calibration parameters, timestamps, platform identifiers, task identifiers, outcome labels, and volunteer-provided ratings. This standardized composition enables training and evaluation across heterogeneous robots, end-effectors, and teleoperation interfaces while maintaining consistent multimodal alignment.

\section{EXPERIMENTS}
\label{sec:experiments}
In this section, we evaluate \method as a training and benchmarking resource for contact-rich industrial manipulation. Our experiments are designed to (i) quantify how well state-of-the-art imitation learning policies leverage the dataset's multimodal signals under high-precision constraints, and (ii) examine how demonstration collection choices affect downstream performance, particularly when the same task is collected via different teleoperation interfaces. Concretely, we train and compare three representative behavior-cloning baselines—ACT \cite{zhao2023learning}, Diffusion Policy (DP) \cite{chi2024diffusionpolicy}, and $\pi_0$ \cite{black2026pi0visionlanguageactionflowmodel} —under consistent training protocols, and report performance using task-relevant metrics such as success rate.

\subsection{Experimental Setup}
\begin{figure}[t]
	\centering
	\includegraphics[width=0.48\textwidth]{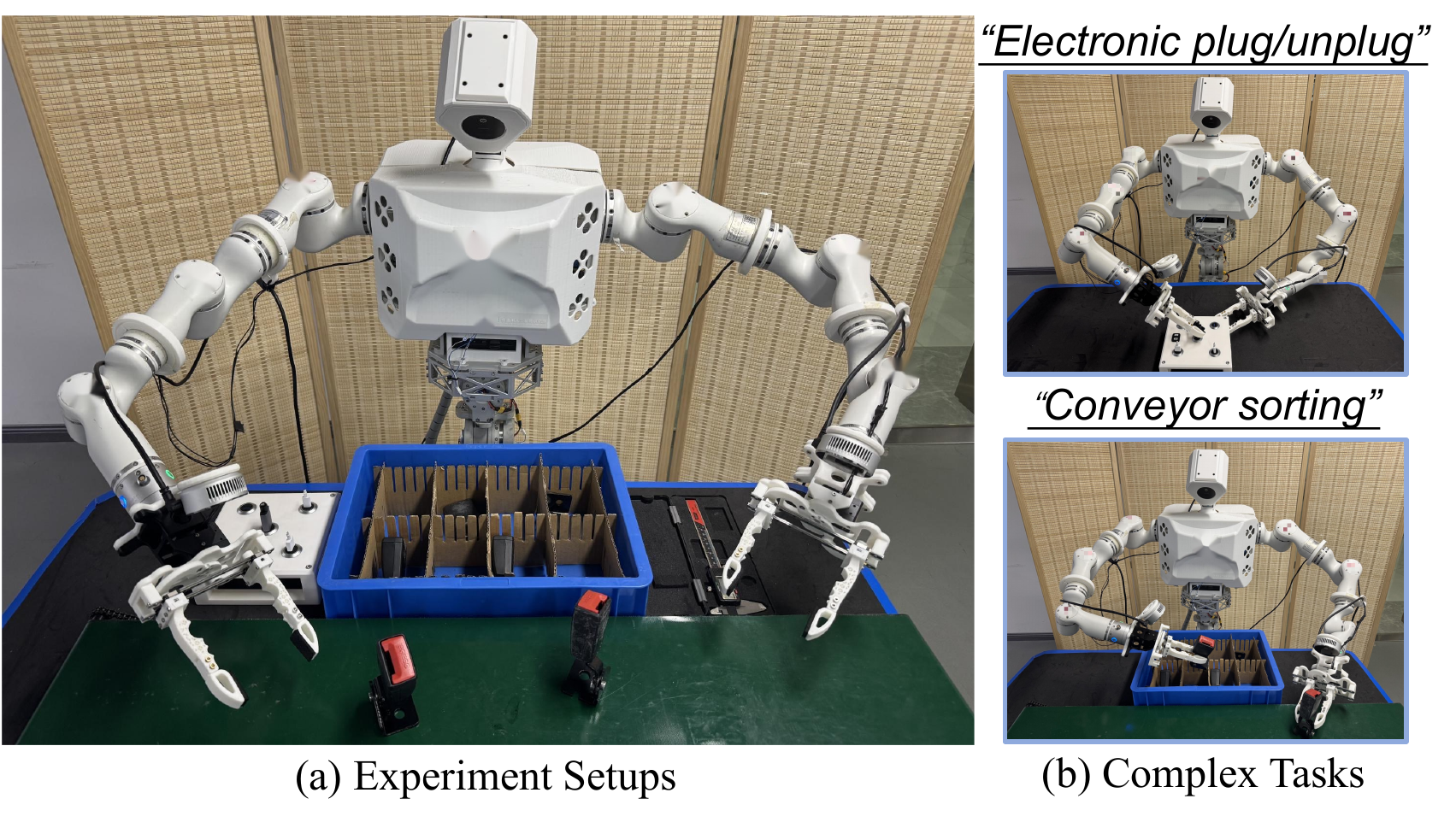}
	\caption{\textbf{Real-World Experiments} with two Realman manipulators.}
	\label{fig:exp}
\end{figure}

\textit{a)\ Platform:} \ 
Experiments are conducted on a bimanual Realman robot with a 3 DoFs waist, matching the embodiment used during data collection. Each arm is equipped with a 3D-printed parallel-jaw gripper, and visual observations are captured using Intel RealSense D515 RGB-D cameras. The workspace layout, background and tabletop settings are kept consistent with the data collection environment to ensure that evaluation reflects the same sensing configuration and physical constraints encountered in the dataset. Fig. \ref{fig:exp} illustrates our robot platform.

\textit{b)\ Procedure:} \ 
We evaluate policies on three representative tasks collected in our industrial workcell: \texttt{electronic component plug/unplug}, \texttt{packaging a vernier caliper} and \texttt{conveyor-based sorting}. These tasks reflect three common categories in industrial manipulation: precise contact-rich operations, product packaging and dynamic object sorting, respectively. For each task, we collect 200 demonstrations, each containing synchronized RGB-D observations, action sequences and 6-axis end-effector force/torque.

Our training follows a two-stage protocol. We first pretrain each model on the full \method dataset (5,000 demonstrations) to learn general multimodal representations and contact-rich control priors. We then fine-tune on the task-specific subset corresponding to the target task. To study data efficiency and the effect of large-scale pretraining, we evaluate: (i) training with 100 vs. 200 task demonstrations, and (ii) with vs. without dataset-level pretraining (e.g., training from scratch using only the task-specific data). All evaluations are performed on the real robot platform and repeated 20 times for each configuration. For the two static tasks (\texttt{electronic component plug/unplug} and \texttt{caliper packaging}), each trial has a 30-second limit, and it is marked successful if the task-specific completion criteria are satisfied within this budget.

\textit{c)\ Implementation Details:} \ 
We convert all collected demonstrations into the LeRobot v3.0 \cite{cadene2024lerobot} dataset format and train policies using the official configuration provided in the LeRobot repository for the three baselines. For each policy, training is split into two stages with a fixed step budget: we allocate $10\%$ of the total optimization steps to pretraining on the full dataset and the remaining $90\%$ to task-specific fine-tuning (e.g., for $\pi_0$ we use 30,000 total steps, with 3,000 steps for pretraining and 27,000 steps for fine-tuning). We set the chunk size to 15, which corresponds to 1 second at 15 Hz. The RGB-D images are scaled to 540 $\times$ 960 during training and testing.

\subsection{Experimental Results}
We present the model's success rates under different numbers of training demonstrations in Table \ref{tab:exp_demo}. Increasing the amount of task-specific data from 100 to 200 demonstrations leads to a clear and consistent performance improvement. This trend is most pronounced for the two static, contact-rich tasks—electronic component plug/unplug and vernier caliper packaging—where additional demonstrations provide better coverage over initial condition variations and, critically, a richer distribution of contact transitions and corrective behaviors. With more demonstrations, policies exhibit (i) higher success rates, (ii) fewer early failures caused by misalignment or premature contact, and (iii) more stable execution trajectories with reduced oscillatory corrections. These findings indicate that even within a fixed task definition, contact-rich industrial manipulation remains data-intensive: collecting a larger set of demonstrations substantially improves the learner's ability to handle small geometric deviations and the long tail of interaction outcomes. For conveyor-based sorting, additional demonstrations also improve performance, but the gains are smaller. This is because sorting on a moving conveyor couples perception latency, prediction error, and time-critical actuation. While more data helps the policy see a broader range of object poses and motion states, success still depends on precise temporal coordination and robust tracking, which are harder to recover purely through behavioral cloning under limited coverage of motion patterns.

\begin{table}[t]
\centering
\small
\begin{tabular}{c cc cc cc}
\toprule
& \multicolumn{2}{c}{\begin{tabular}[c]{@{}c@{}}\texttt{Electronic} \\\texttt{plug/unplug}\end{tabular}}
& \multicolumn{2}{c}{\begin{tabular}[c]{@{}c@{}}\texttt{Calipers} \\\texttt{packaging}\end{tabular}}
& \multicolumn{2}{c}{\begin{tabular}[c]{@{}c@{}}\texttt{Conveyor} \\\texttt{sorting}\end{tabular}} \\
\cmidrule(lr){2-3}
\cmidrule(lr){4-5}
\cmidrule(lr){6-7}
\textbf{Method}
& 100 & 200
& 100 & 200
& 100 & 200\\
\midrule
ACT
& 5 & 10
& 30 & 45
& 15 & 25
\\
\midrule
DP
& 0 & 5
& 25 & 50
& 10 & 25
 \\
\midrule
$\pi_0$
& 10 & \textbf{20}
& 65 & \textbf{80}
& 50 & \textbf{75}
\\
\bottomrule
\end{tabular}
\caption{Average success rates (\%) of three different policies trained with 100 or 200 demonstrations per task and evaluated over 20 episodes.}
\label{tab:exp_demo}
\end{table}

\begin{table}[t]
\centering
\resizebox{\linewidth}{!}{
\begin{tabular}{ccccc}
\toprule
\textbf{Method} & \textbf{Pretrain} & 
 \multicolumn{1}{c}{\begin{tabular}[c]{@{}c@{}}\texttt{Electronic} \\\texttt{plug/unplug}\end{tabular}}
& \multicolumn{1}{c}{\begin{tabular}[c]{@{}c@{}}\texttt{Calipers} \\\texttt{packaging}\end{tabular}}
& \multicolumn{1}{c}{\begin{tabular}[c]{@{}c@{}}\texttt{Conveyor} \\\texttt{sorting}\end{tabular}}\\
\midrule
\multirow{2}{*}{ACT}
& \no & 10 & 45 & 25 \\
& \yes & 10 & 55 & 30 \\
\midrule
\multirow{2}{*}{DP}
& \no & 5 & 50 & 25 \\
& \yes & 10 & 55 & 20 \\
\midrule
\multirow{2}{*}{$\pi_0$}
& \no & 20 & 80 & 75 \\
& \yes & \textbf{25} & \textbf{85} & \textbf{85} \\
\bottomrule
\end{tabular}
}
\caption{Average success rates (\%) of three different policies w/ and w/o pretraining, trained using 200 demonstrations per task and evaluated over 20 episodes.}
\label{tab:exp_pre}
\end{table}

\begin{table}[t]
\centering
\resizebox{\linewidth}{!}{
\begin{tabular}{ccccc}
\toprule
\textbf{Method} & \textbf{Collection} & 
 \multicolumn{1}{c}{\begin{tabular}[c]{@{}c@{}}\texttt{Electronic} \\\texttt{plug/unplug}\end{tabular}}
& \multicolumn{1}{c}{\begin{tabular}[c]{@{}c@{}}\texttt{Calipers} \\\texttt{packaging}\end{tabular}}
& \multicolumn{1}{c}{\begin{tabular}[c]{@{}c@{}}\texttt{Conveyor} \\\texttt{sorting}\end{tabular}}\\
\midrule
\multirow{2}{*}{ACT}
& VR & 5 & 20 & 10 \\
& Exoskeleton & 10 & 45 & 25 \\
\midrule
\multirow{2}{*}{DP}
& VR & 0 & 15 & 20 \\
& Exoskeleton & 5 & 50 & 25 \\
\midrule
\multirow{2}{*}{$\pi_0$}
& VR & 5 & 35 & 40 \\
& Exoskeleton & \textbf{20} & \textbf{80} & \textbf{75} \\
\bottomrule
\end{tabular}
}
\caption{Average success rates (\%) of three different policies trained on 200 demonstrations per task collected via VR vs. Exoskeleton teleoperation, evaluated over 20 episodes.}
\label{tab:exp_tele}
\end{table}

We also illustrate the model's success rate under different training processes in Table \ref{tab:exp_pre}. Pretraining on the full 5,000-trajectory dataset consistently improves downstream performance after fine-tuning on task-specific data. Beyond higher average success rates, pretrained models tend to be more robust: they fail less catastrophically and more often exhibit partial progress or recovery-like behaviors (e.g., re-approach after a minor misalignment, delayed grasp adjustment, or conservative contact engagement). This suggests that large-scale pretraining provides useful priors for (i) multimodal feature extraction, (ii) common motion motifs in industrial workcells, and (iii) contact-rich stabilization strategies that transfer across tasks. In contrast, models trained from scratch on only 100–200 demonstrations are more brittle: they often overfit to dominant trajectories, show less tolerance to perturbations in object pose, and degrade noticeably under minor distribution shifts. The benefit of pretraining is particularly visible when the evaluation involves unexpected disturbances or slight changes in scene configuration. Pretrained policies are more likely to maintain stable behavior under small deviations (e.g., slight object pose shifts, transient occlusions, or contact-induced pose drift). In contrast, policies trained without pretraining often collapse once their execution deviates from the trajectory distribution observed during data collection, leading to error accumulation and task failure.

Despite these improvements, overall performance remains far from satisfactory, especially on the two hardest aspects of our benchmark: dynamic manipulation and precise force-aware operations. For conveyor sorting, failures are often caused by imperfect timing and state estimation under late grasps, unstable grasps due to relative motion at contact, or missed intercepts when object speed/pose deviates from training patterns. For electronic component plug/unplug, failures commonly stem from force-control limitations: slight misalignments lead to jamming, excessive contact forces, or repeated micro-corrections that do not converge within the time limit. These observations indicate that current imitation-learning policies—even with multimodal inputs and large-scale pretraining—still struggle to (i) reason about fast-changing task state in dynamic scenes and (ii) regulate contact forces reliably under tight tolerances.

The above results highlight that more demonstrations and larger scale pretraining help improve coverage and robustness, but they do not fully resolve the underlying difficulty of industrial contact-rich manipulation. It still requires stronger interaction modeling (e.g., explicit contact state estimation), better incorporation of force/visuotactile feedback into closed-loop control, and learning objectives or architectures tailored to time-critical dynamic and precision force regulation.

Beyond the training protocol variations described above, we also study how teleoperation modality influences downstream policy learning. Specifically, we collect demonstrations for the same tasks using VR and an exoskeleton-based platform, with 200 demonstrations per task under each modality, and train policies from scratch under identical model configurations and optimization budgets. The results show that policies trained on VR-collected data consistently achieve lower success rates than those trained on exoskeleton-collected data, and they also exhibit reduced execution efficiency (e.g., longer completion times and more corrective motions). We attribute this gap primarily to differences in perceptual feedback during data collection: in our VR setup, operators observe the scene largely through 2D rendered views, which provide weaker depth and spatial cues than natural 3D perception. Moreover, the exoskeleton platform induces operator motions that more closely match the robot's joint-space kinematics, rather than purely tracking end-effector controller positions as in VR. These limitations can induce systematic imprecision during demonstration—such as suboptimal approach trajectories, noisier alignment behaviors, and less stable contact engagement—thereby degrading demonstration quality and lowering data collection efficiency. These findings suggest that, for high-precision contact-rich industrial manipulation, the choice of teleoperation platform is a critical factor that shapes demonstration fidelity and can materially impact the performance of policies.
\section{Conclusion}
\label{sec:conclusion}
In this paper, we present \method, a large-scale multimodal dataset for contact-rich industrial manipulation under high-precision constraints, covering both NIST Assembly Task Board operations and production-oriented workflows such as packaging and conveyor-based sorting, collected across multiple robot embodiments and end-effectors. The collected data include synchronized RGB-D, visuotactile, joint torques, and proprioception, enabling research on multimodal perception and contact-rich control in realistic industrial settings. We further validate \method by training and evaluating representative imitation-learning baselines on real robots, showing that existing policies' overall performance remains limited for dynamic object manipulation and contact-rich precision operations, highlighting important opportunities for future work on stronger multimodal fusion, interaction-state estimation, and closed-loop contact regulation for generalizable industrial manipulation.
\clearpage

{\normalsize
\bibliographystyle{IEEEtran}
\bibliography{IEEEabrv,main}
}
\end{document}